\documentclass[letterpaper, 10 pt, conference]{ieeeconf}  

\IEEEoverridecommandlockouts                              

\usepackage{nicematrix}
\NiceMatrixOptions{
code-for-first-row = \color{blue} ,
code-for-last-row = \color{blue} ,
code-for-first-col = \color{blue} ,
code-for-last-col = \color{blue}
}

\usepackage{graphics} 
\usepackage{epsfig} 
\usepackage{mathptmx} 
\usepackage{amsmath} 
\usepackage{amssymb}  
\usepackage{multicol}
\usepackage{subcaption}
\usepackage{balance}

\usepackage[bookmarks=true]{hyperref}
\usepackage{comment}
\usepackage{float}	
\usepackage{multirow}
\usepackage{multicol}

\title{\LARGE \bf Dexterous Soft Hands Linearize Feedback-Control \\ for In-Hand Manipulation}

\author{Adrian Sieler$^{1, 2}$ \and Oliver Brock$^{1, 2}$
\thanks{$^{1}$ Robotics and Biology Laboratory, Technische Universität Berlin} 
\thanks{$^{2}$ Science of Intelligence, Research Cluster of Excellence, Berlin}%
\thanks{We gratefully acknowledge funded provided by the Deutsche Forschungsgemeinschaft (DFG, German Research Foundation) under Germany’s Excellence Strategy – EXC 2002/1 “Science of Intelligence” – project number 390523135.}
}

\usepackage{fancyhdr}
\fancypagestyle{IEEECopyright}{
   \lfoot{\copyright2023 IEEE. Accepted at 2023 IEEE/RSJ International Conference on Intelligent Robots and Systems (IROS). \\ DOI: TBA} 
}

\begin{document}

\maketitle
\thispagestyle{IEEECopyright}
\pagestyle{empty}

\begin{abstract}

This paper presents a feedback-control framework for in-hand manipulation (IHM) with dexterous soft hands that enables the acquisition of manipulation skills in the real-world within minutes. We choose the deformation state of the soft hand as the control variable. To control for a desired deformation state, we use coarsely approximated Jacobians of the actuation-deformation dynamics. These Jacobian are obtained via explorative actions. This is enabled by the self-stabilizing properties of compliant hands, which allow us to use linear feedback control in the presence of complex contact dynamics. To evaluate the effectiveness of our approach, we show the generalization capabilities for a learned manipulation skill to variations in object size by $100$~\%, $360$ degree changes in palm inclination and to disabling up to $50$~\% of the involved actuators. In addition, complex manipulations can be obtained by sequencing such feedback-skills.

\end{abstract}

\section{Introduction}\label{sec:intro}

The inherent compliance of soft, dexterous hands contributes importantly to the robustness of manipulation behavior~\cite{adrian-RSS-21, wood_clark, puhlmann_rbo_2022}. Yet, traditional control approaches that require accurate kinematic and/or dynamic models cannot take advantage of this~\cite{sundaralingam_relaxed-rigidity_2019, dlr_impedance}. If object pose or hand configuration are controlled explicitly, as in the standard approaches that are designed to overwrite the system dynamics, the benefits of compliance are ``controlled away.'' This is because the benefits arise exactly from the multitude of task-consistent hand-object dynamics. Here, we develop a novel approach to controlling soft hands, fully leveraging the advantages of compliance.

\begin{figure}[t!]
     \centering
     \begin{subfigure}{0.5\textwidth}
         \centering
         \includegraphics[width=0.9\columnwidth]{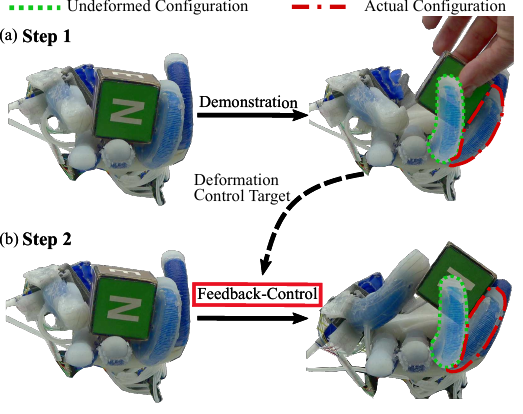}
     \end{subfigure}\\
\caption{An illustration of the complete process of obtaining a feedback-based manipulation primitive for a compliant hand. The desired motion of the object is demonstrated (a). The deformation measurement of the ring finger is recorded at the end of the demonstration. Subsequently, this measurement is used as a control target to replicate the motion using the actuators of the thumb and little finger (b).}
\label{fig:intro_demo}
\end{figure}

\begin{figure}[t!]
     \centering
     \begin{subfigure}{0.5\textwidth}
         \centering
         \includegraphics[width=0.95\columnwidth]{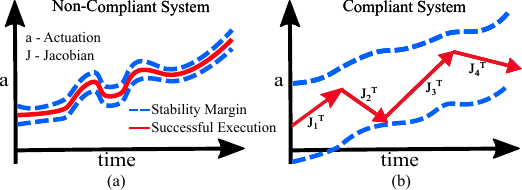}
     \end{subfigure}\\
\caption{Visualization of how compliance enables the linearization of feedback-control for contact-rich manipulation tasks. Details are discussed in Sec.~\ref{sec:intro}.}
\label{fig:intro_two}
\end{figure}

The inherent self-stabilizing properties of soft hands reduce the dependence on accurate control to make progress towards a manipulation goal~\cite{adrian-RSS-21}. This is because the set of \textit{successful} control commands is much larger for compliant than for non-compliant systems. This is illustrated in Fig.~\ref{fig:intro_two} by the width of the area spanned by the blue, dashed lines. A feedback controller based on local linear approximations (Jacobians, red arrows) of the combined hand-object dynamics can perform a multitude of relatively large control adjustments without destabilizing the system (leaving the area between the dashed lines). Jacobians do not need to be perfectly accurate and can be reused even if they are not up-to-date relative to the current hand-object configuration.

\begin{figure}[t!]
    \vspace{2mm}
     \centering
     \begin{subfigure}{0.5\textwidth}
         \centering
         \includegraphics[width=0.88\columnwidth]{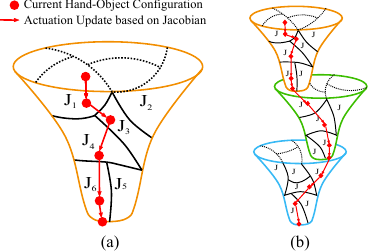}
     \end{subfigure}\\
\caption{A manipulation primitive illustrated as a funnel (a): Jacobians (J), representing local linearizations, tile the volume of the funnel.  Several of these funnels can be sequenced (b) to represent manipulation plans. The focus of this paper lies on obtaining single funnels.}
\label{fig:feedback_funnels}
\end{figure}

In prior work~\cite{adrian-RSS-21}, we demonstrated that in-hand manipulation (IHM) can be structured as a sequence of robust manipulation primitives, metaphorically visualized as manipulation funnels~\cite{mason_mechanics_1985}. Using this metaphor for illustration, we propose a new kind of feedback-funnel~\cite{sequencing_burridge}, defined by a set of Jacobians that are tiling a larger funnel (see Fig.~\ref{fig:feedback_funnels}). This tiling represents a piecewise linear approximation of the nonlinear hand-object dynamics over a large number of successful execution trajectories. Sensor feedback obtained during execution enables the continuous refinement of the funnel's structure, correction of individual tiles, and expansion of the funnel into new parts of the state space. This means that there is no distinction between learning and execution any longer: the controller is built and refined during every successful or unsuccessful execution.

To realize such feedback-funnels, we need a representation of hand-object state that allows taking advantage of compliance. It must accommodate variability in the way the control goal is achieved and it should be easily measurable in spite of the many inherent degrees of freedom of compliant hands. This state must provide rich information about the state of the hand, the object, and the contact forces between them to support diverse manipulation tasks. Furthermore, it must enable the efficient determination of the Jacobians and control commands that lead to reduction of error with respect to a specified control goal, enabling the traversal and refinement of the funnel during execution. We will show in Sec.~\ref{sec:deformation} that the \emph{deformation state} of the hand provides exactly such a representation. Measurements of the soft hands shape can easily be obtained (see Sect.~\ref{sec:implementation_deformation}). Deformation is defined as the difference in these measurements when the object is present and when it is not. 

In Sect.~\ref{sec:jacobian_model_free}, we define a simple notion of continuously-adapting feedback-funnels that are based on Jacobians of the hand-object system. They relate changes in actuation to changes in deformation state. Since deformation can easily be measured, control targets can be extracted from human demonstrations of the desired deformation state: simply move the object the way you want the hand to move it! See Fig.~\ref{fig:intro_demo} for an illustration. The method presented in this paper then autonomously builds a tiled funnel of Jacobians to reach this goal state. We show that the resulting IHM skills generalize for object size and palm inclination; they also remain robust under actuator failures (see Sec.~\ref{sec:experiments}).

The proposed approach to control does not depend on large amounts of data, it does not require physical simulations, it can be computed with negligible computational resources, generates robust and general behavior within seconds based on autonomously gathered real-world experience and does not depend on accurate models of the hand or the world. It can thus be considered a promising alternative to reinforcement learning~\cite{OpenAI, sievers_learning_2022, chen_visual_2022} and model-based approaches~\cite{sundaralingam_relaxed-rigidity_2019, dlr_impedance} to IHM. Interestingly, our approach shares important aspects with biological motor control (see Sec.~\ref{sec:discussion}).

\section{Related Work}

The related work is organized into three main aspects, which are the foundation of the method presented in this work. First, we review the role of deformation in robotic manipulation, which motivates our choice of state. Second, we discuss linear feedback control approaches for deformable systems, because our control approach relies on linearized system dynamics. Third, we summarize existing work on obtaining IHM control goals from demonstration.

\subsection{The Role of Deformation in Robotic Manipulation}

The deformation state of soft actuators interacting with the environment has been used for a variety of different applications, e.g. the computation of contact forces~\cite{deflection_force_kalman, intrinsic_force_sensing}, hand design via the \textit{soft synergy} model~\cite{bicchi_modelling_2011} as well as quantifying and ranking \textit{Morphological Computation} for robot grasping~\cite{mc_good_bad_ugly}. Deformation has also been applied to IHM with a model-based notion of energy fields that are derived from the deformation of the actuators~\cite{energy_morgan}. In comparison, our representation is based on sensor measurements and does not require any kinematic or dynamic models. In addition, recent reinforcement learning approaches to IHM using rigid hands have shown that by incorporating joint-level deformation of an impedance-controller into the state, the agent can learn to ``feel''~\cite{sievers_learning_2022, finger_gaiting_ciocarlie, qi_hand_2022}. Thereby, the dependence on visual feedback is reduced. Overall, the deformation of a compliant system entails a lot of task-relevant information of the combined hand-object system. Therefore, we investigate deformation as the underlying representation to generate IHM behavior with compliant hands.

\subsection{Linearized Feedback-Control of Soft Robots and Objects}

The control of deformable objects~\cite{review_deformable_object} and the control of soft actuators~\cite{review_soft_robot_control} share important properties. Since soft systems are notoriously difficult to model, Jacobian-based methods have been proposed that rely on local linearizations based on measurements of the quantity of interest. For soft actuators \textit{model-less} control approaches have been proposed that rely on an approximation of the Jacobian that relates actuation and actuator-tip-configuration~\cite{yip_model-less_2014, kalman_model_less, soft_arm_model_less}. Similar methods have been applied to control the deformation of a soft object~\cite{navarro_3d_do, data_drive_do, berenson_relaxed_rigidity_do}. Our control algorithm is most closely related to~\cite{yip_model-less_2014}. We generalize this \textit{model-less} control paradigm to systems composed of multiple independent soft actuators that interact with one another through an object held in-hand and apply it to IHM tasks. In addition, we are moving away from controlling the usual end-effector-configuration to a hand-object representation of deformation, as inspired by approaches to deformable object control. Our approach is simpler than previous approaches. We drop the requirement to estimate the Jacobian as accurate as possible, because we keep the same Jacobians even if the hand-object configuration changes. This is enabled by the compliance of the hand.

\subsection{Extracting IHM Control Goals from Demonstration}

Formalizing subgoals for complex IHM tasks to be performed by a robot agent is difficult. Therefore, demonstrations are used to direct behavior generation to regions of the relevant regions of the state-action space. This has been shown for reinforcement learning combined with demonstrated state-action trajectories~\cite{gupta_learning_2016, Rajeswaran-RSS-18}. In~\cite{holo_dex}, visual demonstrations are mapped into a low-dimensional latent space in which a simple Nearest-Neighbor approach is representing the policy. In~\cite{lfd_dmp}, a Dynamic Movement Primitive is learned from a single kinesthetic demonstration to perform within-grasp manipulation. In this work, we only extract the final time step from a demonstrated deformation trajectory that is used as a control target for our Jacobian feedback-controller to reproduce the desired behavior. Our approach is conceptually related to learning from demonstration (LfD) methods based on higher-level intermediate goals~\cite{bernardino_high_level, xing_lfd}.

\section{Deformation---A Hand-Object Representation}
\label{sec:deformation}

\textbf{Deformation:} Soft hands are capable of handling complex contact dynamics by automatically balancing contact forces, dampening impacts, and providing stable grasps through large contact patches. This is enabled by the passive deformation of a soft hand to the shape of the object. The deformation state of the hand can be considered a representation of the combined hand-object system. In general, the deformation is defined as the difference between the hands actual configuration when it is in contact with an object and its configuration during free-motion. Instead of only considering the current position of the hand, the deformation provides information about both the movement of the actuators and additionally the forces that are acting during contact. Furthermore, if multiple actuators are in contact with an object, the change in deformation over time implicitly captures information about the configuration and motion of the object. This renders the deformation an interesting quantity for control.
Representing the deformation of compliant hands comes with some challenges. Sensing the deformation state of every particle a soft hand is composed of is impossible. In addition, modeling the behavior of multiple soft actuators in contact is intractable. Therefore, we will approximate the deformation based on measurements of the shape of the soft actuators that are readily available (see Sec.~\ref{sec:rh3_sensing}).

\textbf{Deformation Approximation:} We consider a single soft actuator with a single degree of actuation $a\in\mathbb{R}$ and assume that we are given a $n$-dimensional sensory modality $s\in\mathbb{R}^n$ that responds to changes of the actuators shape. First, we compute a representation of the actuator's behavior for various actuation states when no contact is present. This is done in a data-driven manner, by densely sampling the actuation-measurement relationship in free-motion. Afterwards, a parameterized model $f_{\Theta_{free}}$ is learned that maps actuation $a$ to measurements $s$: 
\begin{equation}\label{eqn:free_mapping}
    f_{\Theta_{free}}: a \in \mathbb{R} \to s \in \mathbb{R}^n.
\end{equation}
The measurement $s$ depends on $a$ as well as external forces $F_{ext}$ that deform the actuator. Therefore, we can compute an approximation of the deformation state $\Vec{\Delta\,s}$ by evaluating:
\begin{equation}\label{eqn:deformation_delta}
    \Vec{\Delta\,s} = s_t(a, F_{ext}) - f_{\Theta_{free}}(a).
\end{equation}
From now on we will refer to $\Vec{\Delta\,s}$ as the deformation state of a soft actuator. Of course, the chosen sensory modality must allow subtraction and should be proportional to the actual deformation to be applicable to the controller presented in the next section. In addition, we require a low-level actuation scheme that is independent of the compliant interactions of the actuator. Air-mass control~\cite{deimel_mass_control} fulfils this requirement as we will explain in Sec.~\ref{sec:air_mass_control}. In general, this definition of deformation does not impose strict requirements. Therefore, various sensor modalities could be investigated within this framework.

\section{Jacobian-Based Deformation Control}
\label{sec:jacobian_model_free}

In Sec.~\ref{sec:intro}, we discussed how the self-stabilizing properties of soft hands allow us to control the complex hand-object dynamics using coarse local linearizations of these dynamics. In the previous section, we motivated that the deformation state of the hand can be used as a representation of the hand-object system. Here, we derive a deformation controller based on \textit{Deformation Jacobians} that can be reused and updated continuously. In Sec.~\ref{sec:intro}, we referred to this controller as a continuously-adapting feedback-funnel.

For simplicity, we consider a compliant hand consisting of two continuum actuators $A_{1}$ and $A_{2}$, each with one degree of actuation. We assume that both actuators grasp an object in force-closure and that the deformation state (Eq.~\ref{eqn:deformation_delta}) of the two actuators $\Vec{\Delta\,s}_1$ and $\Vec{\Delta\,s}_2$ can be measured in the available sensory modalities. We do not consider any external forces, like gravity. 

Now, we are given a deformation target $\Vec{\Delta\,s}^{target}_1$ in the deformation state of $A_{1}$. This target could have been chosen to modulate the contact force or to move the object within the grasp as described in Sec.~\ref{sec:demonstration}. We can define an error vector based on the current deformation state $\Vec{\Delta\,s}_1$ (Eq. \ref{eqn:deformation_delta}):
\begin{equation}\label{eqn:error}
    \Vec{e} = \Vec{\Delta\,s}_1 - \Vec{\Delta\,s}^{target}_1.
\end{equation}
This error can be minimized in two ways, by changing the actuation $a_{1}$ of the target actuator $A_{1}$ or by changing the actuation $a_{2}$ of the second actuator $A_{2}$. In the following, we derive the actuation update rules for both cases.

\subsection{Derivation of the Deformation Jacobian}\label{sec:derivation_jacobian}

For our two actuator setup, the deformation state of $A_{1}$ takes the following form:
\begin{align}
    \Vec{\Delta\,s}_1 &= s_1(a_{1}, a_{2}) - f_{\Theta_{free}}^1(a_{1}). \label{eqn:fk_sensor_delta}
\end{align}
Note that $a_{2}$ takes the role of the external forces $F_{ext}$ in Eq.~\ref{eqn:deformation_delta}. Therefore, we can change the external forces acting on $A_{1}$ by changing $a_{2}$.
We can now formulate an objective function to be minimized:
\begin{equation}
    C(a_{1}, a_{2}) =  \frac{1}{2}\Vec{e\,}^t \, \Vec{e}. \label{eqn:error_cost}
\end{equation}
First, we compute the partial derivative of $C$ w.r.t. $a_{1}$. The corresponding Jacobian of Eq.~\ref{eqn:fk_sensor_delta} evaluates to:
\begin{equation}\label{eqn:jacobian_target}
    J_{a_{1}} = \frac{\partial s_1}{\partial a_{1}} - \frac{\partial f_{\Theta_{free}}^1}{\partial a_{1}}.
\end{equation}
The partial derivative of $C$ w.r.t. $a_{2}$ takes a similar form but without the $f_{\Theta_{free}}^1$ component.
The corresponding Jacobian of Eq.~\ref{eqn:fk_sensor_delta} evaluates to:
\begin{equation}\label{eqn:jacobian_external}
    J_{a_{2}} = \frac{\partial s_1}{\partial a_{2}}.
\end{equation}
Given the Jacobians defined in Eq.~\ref{eqn:jacobian_target} and Eq.~\ref{eqn:jacobian_external}, the objective function (Eq.~\ref{eqn:error_cost}) can be minimized via gradient descent to update $a_{1}$ and $a_{2}$ with learning rate $\alpha$:
\begin{align} \label{eqn:update_target}
    a_{1}^{new} &= a_{1} - \alpha \, J_{a_{1}}^t \Vec{e} \\ 
    a_{2}^{new} &= a_{2} - \alpha \, J_{a_{2}}^t \Vec{e}.\label{eqn:update_external}
\end{align}
The expression $\frac{\partial f_{\Theta_{free}}^1}{\partial a_{1}}$ can be obtained via auto-differentiation of the learned smooth model $f_{\Theta_{free}}^1$. We assume that no model of $s_1(a_{1}, a_{2})$ is given. Therefore, $\frac{\partial s_1}{\partial a_{1}}$ and  $\frac{\partial s_1}{\partial a_{2}}$ can not be obtained analytically. To approximate these quantities we iteratively execute small actuation changes $\Delta\,a_{1}$ and $\Delta\,a_{2}$ and collect the sensory response. Afterwards, we can compute the Jacobians based on finite differences. This exploratory process is supported by the self-stabilizing properties of the compliant morphology. In general, we can approximate the partial derivative $\frac{\partial s}{\partial a}$ using the following formula:
\begin{align}
\frac{\partial s}{\partial a} \approx \frac{2 s^2 - s^1 - s^3}{2 \Delta\, a}.
\end{align}
Here, $s^1$ represents the initial sensor state before the small actuation change, $s^2$ represents the sensor state after $\Delta\, a$ has been applied, and $s^3$ represents the sensor state after $\Delta\, a$ has been inverted to return to the initial state.

The Jacobians of the deformation states for $A_{1}$ and $A_{2}$ can be computed simultaneously. This allows us to reuse the computed Jacobians in case we want to control for a target specified in the deformation state of $A_{2}$ or in the combined deformation space of both actuators. The full approximated \textit{Deformation Jacobian} $J$ w.r.t. the deformation states of both actuators $\Vec{\Delta\,s}^{full}~=~(\Vec{\Delta\,s}_{1},~\Vec{\Delta\,s}_{2})^t$ takes the form:
\begin{align}
    J = \begin{pNiceMatrix}[first-row, last-col]
             a_1 & a_2 &  \\
             \frac{\partial s_{1}}{\partial a_1} - \frac{\partial f^{1}_{\Theta_{free}}}{\partial a_1} & \frac{\partial s_{1}}{\partial a_2} & \Vec{\Delta s}^{1}\\
            \frac{\partial s_{2}}{\partial a_1} & \frac{\partial s_{2}}{\partial a_2}- \frac{\partial f^{2}_{\Theta_{free}}}{\partial a_2} & \Vec{\Delta s}^{2}
        \end{pNiceMatrix}. \label{eqn:displacement_jacobian}
\end{align}
The \textit{Deformation Jacobian} can be easily extended to multiple actuators with potentially multiple degrees of actuation, which increases the column dimensionality of the Jacobian. Different sensory modalities associated with a single actuator can also be accounted for by capturing these measurements in the finite difference computation. This results in an increase of the row dimensionality of the Jacobian.

\subsection{Updating the Deformation Jacobian}\label{sec:updating_jacobian}

After the actuation values are updated (Eq.~\ref{eqn:update_target},~\ref{eqn:update_external}) based on the relevant entries of the 
\textit{Deformation Jacobian}~(Eq.~\ref{eqn:displacement_jacobian}), the hand-object configuration will have changed. Therefore, the current estimate of $J$ might not be valid anymore. Our primary objective is to achieve our control target, rather than computing $J$ accurately. We apply the same $J$ as long as the task error (Eq.~\ref{eqn:error_cost}) is reduced. If this is not the case anymore, we query a new Jacobian via finite-differences based on explorative actuations. This approach is justified by the self-stabilizing properties of the soft-hand, that enables progress towards the goal despite inaccurate Jacobians, see Fig.~\ref{fig:intro_two}. Nevertheless, we plan to explore different techniques for updating the Jacobian matrix in the future. We can utilize the data generated at each update step, which comes for free. These techniques may include \textit{Broyden's method}~\cite{data_drive_do}, convex optimization~\cite{yip_model-less_2014}, or Kalman Filters~\cite{kalman_model_less}.

\begin{figure}[b!]
    \centering
    \includegraphics[width=0.95\linewidth]{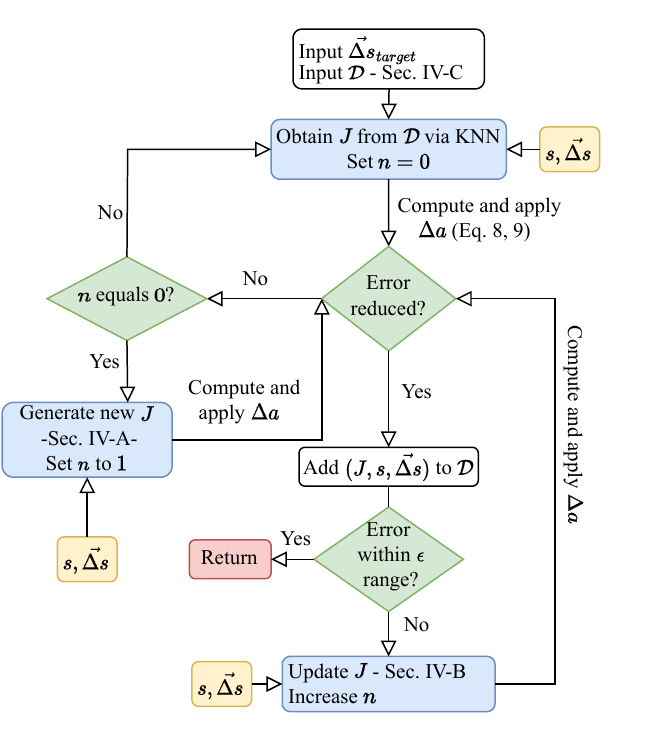}
    \caption{Flowchart of the full Jacobian-based deformation control algorithm. In case no data-set $\mathcal{D}$ is available yet, the algorithm can be executed only with Jacobians that are computed from finite-differences. The generated Jacobians with corresponding sensor and deformation state are used to populate $\mathcal{D}$.}
    \label{fig:flowchart_algorithm}
\end{figure}

\subsection{Storing and Reusing Deformation Jacobians}\label{sec:knn_jacobian}

We want to reuse already computed Jacobians. Whenever a new Jacobian $J_i$ is computed or updated the corresponding absolute sensor state $\mathbf{s}_i$ and deformation state $\Delta\,\Vec{s}_i$ are added to a data-set $\mathcal{D}$ s.t. $\mathcal{D} = (\mathbf{s}_i, \Delta\,\Vec{s}_i, J_i)_{i, \dots, N}$ with $N$ the number of saved tuples. $\mathcal{D}$ represents the Jacobian tiled feedback-funnel, see Fig.~\ref{fig:feedback_funnels}. To reuse the collected Jacobians online, the current absolute sensor state $\mathbf{s}$ and deformation state $\Delta\,\Vec{s}$ are queried and the Nearest-Neighbor (NN) in $\mathcal{D}$ is obtained. The associated Jacobian $J_{NN}$ is then applied to optimize for the current control target. In Fig.~\ref{fig:flowchart_algorithm} the complete algorithm is depicted, which flexibly adds new tuples to $\mathcal{D}$ if the error is not reduced anymore.

\section{Experimental Validation}\label{sec:experiments}

We will structure the experimental section in the following way: First, the robotic platform composed of hand design, low-level control and employed sensory modalities is introduced. Afterwards, we describe how to obtain the deformation state as well as control targets for the available sensory modalities. Based on the derived deformation state, the control and learning pipeline is applied to obtain an in-hand manipulation skill within minutes that requires multiple degrees of actuation and varying contact conditions. In addition, the generalization capabilities of this skill are evaluated. Furthermore, we show how a complex manipulation sequence can be programmed by sequencing various learned feedback-skills. Finally, we apply deformation control to enable table-top sliding of various objects.

\subsection{Sensorized RBO Hand 3}\label{sec:rh3_sensing}

\begin{figure}[h!]
    \centering
    \includegraphics[width=0.95\linewidth]{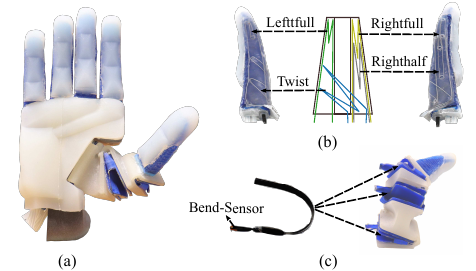}
    \caption{The \textit{RBO Hand 3} (a) with corresponding four-dimensional strain sensor layout on a two-compartment finger (b) and bend-sensor placement on thumb-scaffold (c).}
    \label{fig:rh3_sensors}
\end{figure}

The \textit{RBO Hand 3} (RH3)~\cite{puhlmann_rbo_2022} is an anthropomorphic, soft robotic hand. In total, the hand consists of 16 independent pneumatic actuators. The fingers are made of silicon and have two air chambers each, while the thumb-tip is a single-chamber actuator. In addition, the hand has an opposable thumb mechanism with three joints that are moved via soft bellows actuators made of coated nylon fabric. Another bellow actuator is placed inside the palm for better opposition of thumb and little finger. Three more bellows are between the fingers for abduction and adduction, but these are not used in this work.

We extend the RH3 with two different sensory modalities. First, liquid metal strain sensors~\cite{wall_strain} wrapped around the silicon fingers. Second, commercially available bend sensors (1-Axis \textit{Bendlabs} Soft Flex Sensor) placed in the compliant joints of the thumb, see Fig.~\ref{fig:rh3_sensors}.

\subsection{Equilibrium Point Control Through Air-Mass Control}
\label{sec:air_mass_control}

We control our soft actuators~\cite{puhlmann_rbo_2022} by changing the enclosed air-mass in each actuator~\cite{deimel_mass_control}. This low-level control scheme is of great importance for our proposed approach. As mentioned in Sec.~\ref{sec:deformation}, to compute the deformation state (Eq.~\ref{eqn:deformation_delta}) we need an actuation quantity that is not influenced by the compliant interactions of the hand with the environment. Air-mass as opposed to commonly used quantities for control like pressure, position or force fulfills this requirement. The air-mass enclosed in an actuator uniquely defines its equilibrium position that it would attain if no external forces are present.

\subsection{Implementation of the Deformation State}
\label{sec:implementation_deformation}

We consider a two-compartment pneuflex actuator augmented with a strain sensor layout consisting of four sensors, see Fig.~\ref{fig:rh3_sensors}b. First, the maximum air-mass for each compartment is specified by increasing the air-mass in free-motion until $250$ kPa are reached. At this pressure value, the actuator movement is maximum. Afterwards, we iterate through combinations of the two air-mass values for both compartments to compute the free-motion mapping $f_{\Theta_{free}}$ (Eq.~\ref{eqn:free_mapping}) for the four strain sensors. This mapping is represented as a two-layer neural network with $5$ and $3$ neurons and $tanh$ activation functions. Now, the deformation state can simply be computed by reading the current strain sensor value and subtracting $f_{\Theta_{free}}$ evaluated at the current actuation state. The same procedure is applied to compute the deformation state of the bend sensors for the three joints of the thumb-scaffold, see Fig.~\ref{fig:rh3_sensors}c.

\subsection{Obtaining IHM Control Targets Through Demonstration}\label{sec:demonstration}

We assume that the object that should be manipulated is already in contact with the soft hand. Now, the operator can demonstrate a desired object motion by manually moving the object. This process deforms the actuators which is recognized by the sensors. Throughout the demonstration the actuation values of the actuators remain fixed and the available deformation state (Eq.~\ref{eqn:deformation_delta}) is collected. Afterwards, the operator needs to decide for the deformation dimensions that should represent the manipulation goal. If this is decided, the relevant entries are selected from the last time-step of the demonstrated deformation trajectory. Finally, the demonstrator chooses a set of actuators that should be adjusted to reach the desired control target. The actuators should be selected in a way s.t. they are able to replicate the external forces exerted by the operator during the demonstration, thereby reproducing the motion of the object.
 
\subsection{Rapid Learning of a Robust Manipulation Skill}
\label{sec:robust_skill}

In this section, we demonstrate the application of the Jacobian-based deformation control framework presented in Sec.~\ref{sec:jacobian_model_free} and evaluate the continuous adaptation capabilities of the system to changes in the task. First, the hand is inflated to an initial position depicted in Fig.~\ref{fig:intro_demo}a. From this position a demonstration to shift the object towards the little finger is provided, see Fig.~\ref{fig:intro_demo}b. The control target is represented in the four-dimensional strain deformation state of the ring finger. A total of six degrees of actuation are controlled to reach the task, four related to the thumb and two related to the little finger. The controller runs at $5$~Hz. The first Jacobians are collected for a cube of size $4.5$~cm. All sensory dimensions are normalized using empirically determined minimum and maximum values. We apply the automatic learning rate computation based on the current error $e$ as presented in~\cite{jacobian_methods} for Jacobian transposed inverse-kinematics. Three Jacobian/measurement pairs are collected. The measurements consist of absolute and deformation data for T2, T3 bend-sensors as well as ring and thumb strain sensors (Video: \url{https://youtu.be/nTtkNW59dgk}).

The subsequent generalization experiments always start off with the stored data (Sec.~\ref{sec:knn_jacobian}) and new Jacobians are obtained if no progress towards the goal is obtained. We terminate the execution after five different Jacobians have been applied to reach the goal.

\begin{figure}[t!]
     \vspace{2mm}
     \centering
     \begin{subfigure}{0.49\textwidth}
         \centering
         \includegraphics[width=0.95\columnwidth]{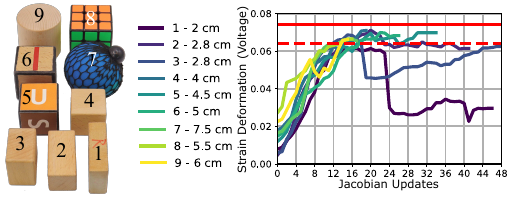}
         \caption{Results Object Variation Experiment}
         \label{fig:object_variation}
     \end{subfigure}\\
     \vspace{2mm}
     \begin{subfigure}{0.49\textwidth}
         \centering
         \includegraphics[width=0.95\columnwidth]{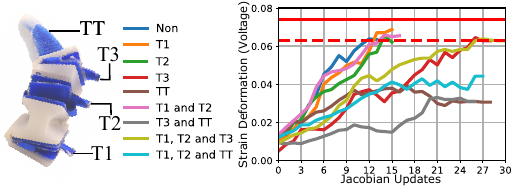}
         \caption{Results Disabled Actuators Experiment}
         \label{fig:disabled_actuators}
     \end{subfigure}\\
     \vspace{2mm}
     \begin{subfigure}{0.49\textwidth}
         \centering
         \includegraphics[width=0.95\columnwidth]{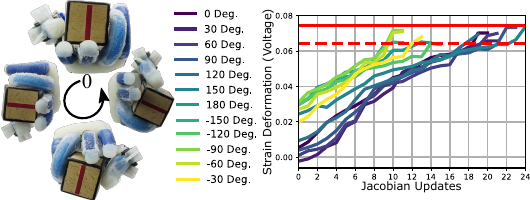}
         \caption{Results Palm Inclination Experiment}
         \label{fig:palm_inclinations}
     \end{subfigure}\\
\caption{Generalization of the shift skill to variations in object size by $100$ \%, $360$ degree changes in palm inclination and to disabling up
to $50$ \% of the involved actuators. All plots show the evolution of the most variable strain deformation dimension (twist, see Fig.~\ref{fig:rh3_sensors}). The red line indicates the control target while the dotted red line indicates the threshold for a successful execution. The variation the deformation at the beginning could be interpreted as being at the funnel entrance, while ending up in the area between target and threshold indicates the funnel exit, see Fig.~\ref{fig:feedback_funnels}}.
\label{fig:variation_experiments_skill}
\end{figure}

\textbf{Object Properties:} The manipulation skill learned for one object can be applied successfully to objects ranging in width from $2.8$ cm to $6$ cm, as shown in Fig.~\ref{fig:object_variation}, which illustrates the deformation error over time per iteration. If the object width decreases, the thumb may slip, but a new Jacobian can account for the change and allow the system to recover, as seen with object $2$ in Fig.~\ref{fig:object_variation}. (Video: \url{https://youtu.be/UWABG_5sLT8})

\textbf{Disabled Actuators:} This experiment simulates a scenario where an actuator can no longer be inflated. In response, the controller can acquire new Jacobians to deal with the change of available degrees of actuation. Fig.~\ref{fig:disabled_actuators} depicts the deformation evolution towards the goal. The results indicate that the thumb-tip (TT) is the most important actuator for this skill. (Video: \url{https://youtu.be/HuL7M4LQwWE})

\textbf{Palm Inclination:} The controller can handle changes in palm inclination by $360$ degree, see Fig.~\ref{fig:palm_inclinations}. If gravity points towards the ring finger, the controller needs to perform less adjustments because gravity supports the motion. When gravity points in the opposite direction, more Jacobian updates are required. This is apparent in the two clusters that form in Fig.~\ref{fig:palm_inclinations} (Video: \url{https://youtu.be/ZR6H8GZocbc}). The results motivate to consider wrist motions as additional degrees of actuation to be integrated into the \textit{Deformation Jacobians}. 

\subsection{Cube Reconfiguration by Sequencing Feedback-Funnels}

\begin{figure}[h!]
    \centering
    \includegraphics[width=0.95\linewidth]{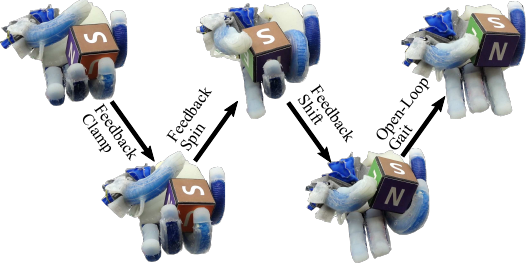}
    \caption{Sequencing of three feedback skills and one open-loop skill to clamp, rotate and translate a cube.}
    \label{fig:sequencing}
\end{figure}

To showcase the capabilities of our proposed approach, we teach the RH3 three feedback-skills to shift and rotate a cube. We program the manipulation skills by demonstration as presented in Sec.~\ref{sec:demonstration}. The following skills with corresponding effect are obtained: Clamp (Squeezing the cube between thumb and middle and ring finger), Spin (Counter-clockwise rotation by $90$ degree), Shift (Translate object towards little finger). In addition, we add one open-loop skills to gait the object over to the little finger by simply deflating both compartments of the ring-finger. The sequence of the skills is specified offline, see Fig.~\ref{fig:sequencing}. In the following video, the full skill acquisition and execution of the sequenced feedback-primitives for cubes of different size are presented:~\url{https://youtu.be/Y7Pl_ZCV1bo}.

\subsection{Table-Top Object Sliding Through Deformation Control}

\begin{figure}[h!]
    \centering
    \includegraphics[width=0.95\linewidth]{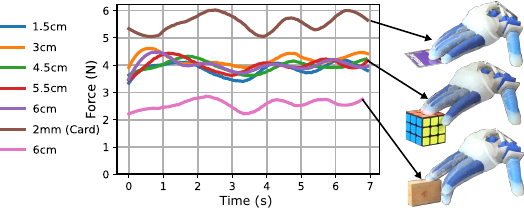}
    \caption{Generated force between RBO Hand 3 and object while sliding objects of different heights $50$ cm along a table. The force generated through the deformation of the soft hand is measured with a \textit{Schunk FTN-Mini-40} mounted between hand and arm. The amount of force applied depends on how many and how fingers are deformed when getting into contact with the object and potentially the table. The force remains within a margin of $1$N.}
    \label{fig:force_sliding}
\end{figure}

Our method is not specifically tailored to in-hand manipulation tasks but can also be applied to any contact-rich manipulation task in which a deformation state of a soft manipulator can be identified that characterizes the required hand-object interactions to successfully complete the task. In this experiment, we control the downward and upward motion of a \textit{Franka Panda} to achieve and maintain a desired hand deformation, allowing for sliding an object across a table.
First, the desired deformation of the ring-finger is demonstrated (Sec.~\ref{sec:demonstration}). Second, a single Jacobian that represents the effect of Panda motion on the ring-finger deformation is estimated and applied to reach the target deformation (Sec.~\ref{sec:jacobian_model_free}). Third, a predefined up- and sideward trajectory is executed once the desired deformation state is reached, potentially causing the hand to lose contact with the object. However, using the deformation feedback and the estimated Jacobian, the trajectory is adjusted online to maintain a constant hand-object interaction. This enables robust sliding of various objects across a table (Video: \url{https://youtu.be/H6BvXXB-mZ8}). This is by far not a general solution for sliding objects across a wide range of different surfaces, but rather a demonstration of the potential applications of our simple and computational cheap approach. The compliant hand's deformation state implicitly captures the total contact force between hand and environment~-~Fig.~\ref{fig:force_sliding}.

\section{Limitations}

The Jacobian acquisition based on finite differences can fail if the actuators start to slip, as explorative motions can accelerate this effect. In addition, the deformation dimensions that the controller should optimize for need to be specified manually. The same applies for the actuators that should be controlled to reach the goal. Also, the acquired control target through demonstration might not be reachable by controlling the actuators. This requires tuning of the success threshold in the context of the skills that should be executed afterwards. Again, the intrinsic compliance of soft hands reduces the need for precisely reaching the control goal to proceed, as demonstrated in~\cite{Morgan} where the concept of inflated switching regions is introduced.
The controller requires the actuators of interest to be in contact to obtain deformation feedback. Therefore, the initial configuration of the hand-object system still needs to be manually designed. This motivates investigation of planning algorithms to reason about possibilities to release and establish contact after a control target has been reached.

\section{Similarity to Biological Control}\label{sec:discussion}

Our approach shares conceptual similarities to the \textit{Equilibrium Point (EP) Hypothesis} also called \textit{Threshold Control Theory (TCT)}~\cite{ep_hypothesis}, an established theory for biological motor control. The EP Hypothesis proposes that the equilibrium configuration of the arm/hand is the control variable modulated by the central nervous system as opposed to the actual torques/forces that are generated in the interaction. For our synthetic system we realize this with air-mass control, see Sec.~\ref{sec:air_mass_control}. Furthermore, the \textit{EP Hypothesis} relies on the perception of the difference between equilibrium configuration and the actual configuration of the body to reason about and control contact with the environment. We call this deformation state, see Sec.~\ref{sec:deformation}. The EP Hypothesis suggests that the body's natural dynamics can take over predictive and anticipatory functions in this process. Similarly, we can interpret the self-stabilizing properties of soft hands as computational resources of the body that can simplify the control of the deformation state via a linear feedback-controller, see Fig.~\ref{fig:intro_two}.

The concept of using roughly approximated Jacobians for control also bears similarity to another theory in motor control~\cite{sloppy}. In this work, the authors propose that humans solve control problems with inaccurate Jacobians of the relevant target quantities. In our approach, we achieve robust manipulation behavior with \textit{good enough} Jacobians because the compliance of the hand can account for the remaining uncertainty. The ”Act on the most nimble” (AMN) rule presented in~\cite{sloppy} offers interesting ideas for stabilizing rapidly changing sub-spaces in the deformation space, which could be used to update the Jacobian online to prevent slippage.

Overall, the observed biological similarities provide us with interesting algorithmic ideas on how to effectively use the natural dynamics of the body for robotic manipulation. This could potentially allow robots to come closer to humans in terms of data efficiency when refining and acquiring new skills in the real world

\section{Conclusion}

We have introduced a feedback-control framework for soft hands that takes full advantage of their compliance. We represent the combined hand-object system implicitly in the deformation state of the soft hand. This definition of state is well-suited to leverage the self-stabilizing properties of soft hands, which enable control of the complex hand-object dynamics by locally linearizing the deformation dynamics. Our approach can generate in-hand manipulation skills in the real world within minutes. These skills can be interpreted as feedback-funnels, which are defined by a set of \textit{Deformation Jacobians}. The skills can be continuously adapted by querying new Jacobians on the fly, which refines the feedback-funnel. This property allows for strong generalization to changing object properties, palm inclinations, and disabled actuators. We hope our findings will inspire further research into how compliance can be used to develop data-efficient methods for obtaining generalizable and reusable in-hand manipulation behavior in the real world.


\bibliographystyle{IEEEtran}
\balance
\bibliography{references}

\begin{thebibliography}{10}
\providecommand{\url}[1]{#1}
\csname url@rmstyle\endcsname
\providecommand{\newblock}{\relax}
\providecommand{\bibinfo}[2]{#2}
\providecommand\BIBentrySTDinterwordspacing{\spaceskip=0pt\relax}
\providecommand\BIBentryALTinterwordstretchfactor{4}
\providecommand\BIBentryALTinterwordspacing{\spaceskip=\fontdimen2\font plus
\BIBentryALTinterwordstretchfactor\fontdimen3\font minus
  \fontdimen4\font\relax}
\providecommand\BIBforeignlanguage[2]{{%
\expandafter\ifx\csname l@#1\endcsname\relax
\typeout{** WARNING: IEEEtran.bst: No hyphenation pattern has been}%
\typeout{** loaded for the language `#1'. Using the pattern for}%
\typeout{** the default language instead.}%
\else
\language=\csname l@#1\endcsname
\fi
#2}}

\bibitem{adrian-RSS-21}
A.~Bhatt, A.~Sieler, S.~Puhlmann, and O.~Brock, ``{Surprisingly Robust In-Hand
  Manipulation: An Empirical Study},'' in \emph{Proceedings of Robotics:
  Science and Systems}, 2021.

\bibitem{wood_clark}
S.~Abondance, C.~B. Teeple, and R.~J. Wood, ``A dexterous soft robotic hand for
  delicate in-hand manipulation,'' \emph{Robotics and Automation Letters},
  vol.~5, no.~4, pp. 5502--5509, 2020.

\bibitem{puhlmann_rbo_2022}
S.~Puhlmann, J.~Harris, and O.~Brock, ``{RBO} hand 3: A platform for soft
  dexterous manipulation,'' \emph{Transactions on Robotics}, vol.~38, no.~6,
  pp. 3434--3449, 2022.

\bibitem{sundaralingam_relaxed-rigidity_2019}
B.~Sundaralingam and T.~Hermans, ``Relaxed-rigidity constraints: kinematic
  trajectory optimization and collision avoidance for in-grasp manipulation,''
  \emph{Autonomous Robots}, vol.~43, no.~2, pp. 469--483, 2019.

\bibitem{dlr_impedance}
M.~Pfanne, M.~Chalon, F.~Stulp, H.~Ritter, and A.~Albu-Schäffer,
  ``Object-level impedance control for dexterous in-hand manipulation,''
  \emph{IEEE Robotics and Automation Letters}, vol.~5, no.~2, pp. 2987--2994,
  2020.

\bibitem{mason_mechanics_1985}
M.~T. {Mason}, ``The mechanics of manipulation,'' in \emph{International
  Conference on Robotics and Automation (ICRA)}, vol.~2, 1985, pp. 544--548.

\bibitem{sequencing_burridge}
R.~R. Burridge, A.~A. Rizzi, and D.~E. Koditschek, ``Sequential composition of
  dynamically dexterous robot behaviors,'' \emph{The International Journal of
  Robotics Research}, vol.~18, no.~6, pp. 534--555, 1999.

\bibitem{OpenAI}
O.~M. Andrychowicz, B.~Baker, M.~Chociej, R.~Jozefowicz, B.~McGrew,
  J.~Pachocki, A.~Petron, M.~Plappert, G.~Powell, A.~Ray, \emph{et~al.},
  ``Learning dexterous in-hand manipulation,'' \emph{The International Journal
  of Robotics Research}, vol.~39, no.~1, pp. 3--20, 2020.

\bibitem{sievers_learning_2022}
L.~Sievers, J.~Pitz, and B.~Bauml, ``Learning purely tactile in-hand
  manipulation with a torque-controlled hand,'' in \emph{International
  Conference on Robotics and Automation ({ICRA})}, 2022, pp. 2745--2751.

\bibitem{chen_visual_2022}
\BIBentryALTinterwordspacing
T.~Chen, M.~Tippur, S.~Wu, V.~Kumar, E.~Adelson, and P.~Agrawal, ``Visual
  dexterity: In-hand dexterous manipulation from depth,'' 2022. [Online].
  Available: \url{http://arxiv.org/abs/2211.11744}
\BIBentrySTDinterwordspacing

\bibitem{deflection_force_kalman}
D.~C. Rucker and R.~J. Webster, ``Deflection-based force sensing for continuum
  robots: A probabilistic approach,'' in \emph{International Conference on
  Intelligent Robots and Systems (IROS)}, 2011, pp. 3764--3769.

\bibitem{intrinsic_force_sensing}
K.~Xu and N.~Simaan, ``An investigation of the intrinsic force sensing
  capabilities of continuum robots,'' \emph{Transactions on Robotics}, vol.~24,
  no.~3, pp. 576--587, 2008.

\bibitem{bicchi_modelling_2011}
A.~Bicchi, M.~Gabiccini, and M.~Santello, ``Modelling natural and artificial
  hands with synergies,'' \emph{Philosophical Transactions of the Royal Society
  B: Biological Sciences}, vol. 366, no. 1581, pp. 3153--3161, 2011.

\bibitem{mc_good_bad_ugly}
K.~Ghazi-Zahedi, R.~Deimel, G.~Montúfar, V.~Wall, and O.~Brock,
  ``Morphological computation: The good, the bad, and the ugly,'' in
  \emph{International Conference on Intelligent Robots and Systems (IROS)},
  2017, pp. 464--469.

\bibitem{energy_morgan}
R.~R. Ma, W.~G. Bircher, and A.~M. Dollar, ``Toward robust, whole-hand caging
  manipulation with underactuated hands,'' in \emph{International Conference on
  Robotics and Automation (ICRA)}, 2017, pp. 1336--1342.

\bibitem{finger_gaiting_ciocarlie}
G.~Khandate, M.~Haas-Heger, and M.~Ciocarlie, ``On the feasibility of learning
  finger-gaiting in-hand manipulation with intrinsic sensing,'' in
  \emph{International Conference on Robotics and Automation (ICRA)}, 2022, pp.
  2752--2758.

\bibitem{qi_hand_2022}
\BIBentryALTinterwordspacing
H.~Qi, A.~Kumar, R.~Calandra, Y.~Ma, and J.~Malik, ``In-hand object rotation
  via rapid motor adaptation.'' [Online]. Available:
  \url{http://arxiv.org/abs/2210.04887}
\BIBentrySTDinterwordspacing

\bibitem{review_deformable_object}
J.~Zhu, A.~Cherubini, C.~Dune, D.~Navarro-Alarcon, F.~Alambeigi, D.~Berenson,
  F.~Ficuciello, K.~Harada, J.~Kober, X.~Li, J.~Pan, W.~Yuan, and M.~Gienger,
  ``Challenges and outlook in robotic manipulation of deformable objects,''
  \emph{Robotics and Automation Magazine}, vol.~29, no.~3, pp. 67--77, 2022.

\bibitem{review_soft_robot_control}
T.~George~Thuruthel, Y.~Ansari, E.~Falotico, and C.~Laschi, ``Control
  strategies for soft robotic manipulators: A survey,'' \emph{Soft Robotics},
  vol.~5, no.~2, pp. 149--163, 2018.

\bibitem{yip_model-less_2014}
M.~C. Yip and D.~B. Camarillo, ``Model-less feedback control of continuum
  manipulators in constrained environments,'' \emph{Transactions on Robotics},
  vol.~30, no.~4, pp. 880--889, 2014.

\bibitem{kalman_model_less}
M.~Li, R.~Kang, D.~T. Branson, and J.~S. Dai, ``Model-free control for
  continuum robots based on an adaptive kalman filter,'' \emph{Transactions on
  Mechatronics}, vol.~23, no.~1, pp. 286--297, 2018.

\bibitem{soft_arm_model_less}
Y.~Jin, Y.~Wang, X.~Chen, Z.~Wang, X.~Liu, H.~Jiang, and X.~Chen, ``Model-less
  feedback control for soft manipulators,'' in \emph{International Conference
  on Intelligent Robots and Systems (IROS)}, 2017, pp. 2916--2922.

\bibitem{navarro_3d_do}
D.~Navarro-Alarcon, H.~M. Yip, Z.~Wang, Y.-H. Liu, F.~Zhong, T.~Zhang, and
  P.~Li, ``Automatic 3-d manipulation of soft objects by robotic arms with an
  adaptive deformation model,'' \emph{Transactions on Robotics}, vol.~32,
  no.~2, pp. 429--441, 2016.

\bibitem{data_drive_do}
F.~Alambeigi, Z.~Wang, R.~Hegeman, Y.-H. Liu, and M.~Armand, ``A robust
  data-driven approach for online learning and manipulation of unmodeled 3-d
  heterogeneous compliant objects,'' \emph{Robotics and Automation Letters},
  vol.~3, no.~4, pp. 4140--4147, 2018.

\bibitem{berenson_relaxed_rigidity_do}
D.~Berenson, ``Manipulation of deformable objects without modeling and
  simulating deformation,'' in \emph{International Conference on Intelligent
  Robots and Systems (IROS)}, 2013, pp. 4525--4532.

\bibitem{gupta_learning_2016}
A.~Gupta, C.~Eppner, S.~Levine, and P.~Abbeel, ``Learning dexterous
  manipulation for a soft robotic hand from human demonstrations,'' in
  \emph{International Conference on Intelligent Robots and Systems ({IROS})},
  2016, pp. 3786--3793.

\bibitem{Rajeswaran-RSS-18}
A.~Rajeswaran, V.~Kumar, A.~Gupta, G.~Vezzani, J.~Schulman, E.~Todorov, and
  S.~Levine, ``Learning complex dexterous manipulation with deep reinforcement
  learning and demonstrations,'' in \emph{Proceedings of Robotics: Science and
  Systems}, Pittsburgh, Pennsylvania, June 2018.

\bibitem{holo_dex}
\BIBentryALTinterwordspacing
S.~P. Arunachalam, I.~Güzey, S.~Chintala, and L.~Pinto, ``Holo-dex: Teaching
  dexterity with immersive mixed reality,'' 2022. [Online]. Available:
  \url{https://arxiv.org/abs/2210.06463}
\BIBentrySTDinterwordspacing

\bibitem{lfd_dmp}
G.~Solak and L.~Jamone, ``Learning by demonstration and robust control of
  dexterous in-hand robotic manipulation skills,'' in \emph{International
  Conference on Intelligent Robots and Systems (IROS)}, 2019, pp. 8246--8251.

\bibitem{bernardino_high_level}
U.~Prieur, V.~Perdereau, and A.~Bernardino, ``Modeling and planning high-level
  in-hand manipulation actions from human knowledge and active learning from
  demonstration,'' in \emph{International Conference on Intelligent Robots and
  Systems (IROS)}, 2012, pp. 1330--1336.

\bibitem{xing_lfd}
X.~Li and O.~Brock, ``Learning from demonstration based on environmental
  constraints,'' \emph{Robotics and Automation Letters}, vol.~7, no.~4, pp.
  10\,938--10\,945, 2022.

\bibitem{deimel_mass_control}
R.~Deimel, M.~Radke, and O.~Brock, ``Mass control of pneumatic soft continuum
  actuators with commodity components,'' in \emph{International Conference on
  Intelligent Robots and Systems (IROS)}, 2016, pp. 774--779.

\bibitem{wall_strain}
V.~Wall, G.~Zöller, and O.~Brock, ``A method for sensorizing soft actuators
  and its application to the rbo hand 2,'' in \emph{IEEE International
  Conference on Robotics and Automation (ICRA)}, 2017, pp. 4965--4970.

\bibitem{jacobian_methods}
S.~Buss, ``Introduction to inverse kinematics with jacobian transpose,
  pseudoinverse and damped least squares methods,'' \emph{Transactions in
  Robotics and Automation}, vol.~17, 2004.

\bibitem{Morgan}
A.~S. Morgan, K.~Hang, B.~Wen, K.~Bekris, and A.~M. Dollar, ``Complex in-hand
  manipulation via compliance-enabled finger gaiting and multi-modal
  planning,'' \emph{Robotics and Automation Letters}, vol.~7, no.~2, pp.
  4821--4828, 2022.

\bibitem{ep_hypothesis}
A.~Feldman and M.~Levin, ``The equilibrium-point hypothesis – past, present
  and future,'' \emph{Advances in experimental medicine and biology}, vol. 629,
  pp. 699--726, 2009.

\bibitem{sloppy}
V.~M. Akulin, F.~Carlier, S.~Solnik, and M.~L. Latash, ``Sloppy, but
  acceptable, control of biological movement: Algorithm-based stabilization of
  subspaces in abundant spaces,'' \emph{Journal of Human Kinetics}, vol.~67,
  no.~1, pp. 49--72, 2019.

\end{thebibliography}

\end{document}